# The role of behavior modifiers in representation development

**Carlos R. de la Mora B.**
AI-Lab, Vrije Universiteit Brussel
Building G 10 room 725. Pleinlaan 2. 1050 Brussels – Belgium
carlos@arti.vub.ac.be

**Carlos Gershenson**
CLEA, Vrije Universiteit Brussel
Krijgskundestraat 33
B-1160 Brussels - Belgium
cgershen@vub.ac.be

**Angélica García-Vega**
MIA, Universidad Veracruzana
Sebastián Camacho 5. CP 91100
Xalapa, Veracruz - México
angegarcia@uv.mx

**Abstract**

We address the problem of the development of representations and their relationship to the environment. We study a software agent which develops in a network a representation of its simple environment which captures and integrates the relationships between agent and environment through a *closure mechanism*. The inclusion of a variable behavior modifier allows better representation development. This can be confirmed with an *internal* description of the *closure mechanism*, and with an *external* description of the properties of the representation network.

**Keywords**: Closure, representation development, behavior modifiers, emotion, affective states, biological motivations, symbol-matter problem.

## 1. Introduction

Ziemke (**2001**) distinguished five distinct categories of embodiment relevant to the epigenetic phenomenon: 1) Structural coupling between the agent and its environment; 2) Historical embodiment as the result of a history of structural coupling. This includes the first category and is so general that a rock can be considered as having historical or structural embodiment; 3) Physical embodiment being systems connected with the world via a set of sensors and actuators, to hold the "Physical Grounding Hypothesis" (**Brooks, 1990**); 4) "Organismoid" embodiment or organism-like bodily form, stressing the specific restrictions related to specific cognitive problems. 5) "Organismic" embodiment of autopoietic living systems, based on the idea of Maturana and Varela (**1987**) and von Uexküll (**1928**) that cognition is what living systems do in *interaction* with their environment, reflecting that "there is a clear difference between living organisms, which are autonomous and autopoietic, and man-made machines, which are heteronomous and allopoietic". In this perspective, only with organismic embodiment could produce artificial autopoietic systems.

However, organismic embodiment includes development in all the biological ways of organization, since living systems *develop* in time and in an integrated way along the POE space (with phylogenic, ontogenetic and epigenetic organizations as dimensions) (**Sipper et al. 1997**) when they interact with their environment. Within this perspective, classical specific-task artificial systems are *points* in the POE space, showing dynamics specific-to-contexts. On the other hand, biological systems trace *trajectories* in the POE space. Therefore, development research should be interested in processes which allow trajectories, although practical difficulties (**Ibid**), take us to consider simplifications, generally constraining ourselves to the epigenetic or phylogenic dimension.

One way to follow is to consider the degree of intelligence of a system related to its use and creation of representations (**Steels, 1995**). Representations can be explicit (or symbolic) but also implicit (or emergent (**Steels, 2003**)).

A necessary property of an epigenetic robot is the ability to generate representations (**Zlatev and Balkenius, 2001**). However, it is not clear nor very well understood what internal representations *are*, or how they develop (external representations require an internal representation to acquire meaning). It seems it is a common assumption to think of internal representations as things, or *material* objects, as sketched in Figure 1a. Under this assumption, if we dissect the brain of an animal, we would find representations somewhere there. Contrasting this assumption, we believe that internal representations are the intertwined *relationships* between significant perceptions and significant actuations of an embodied agent interacting in an environment, as portrayed in Figure 1b. Representations do have a physical structure, but this is only a part of them. This structure is manipulated by a mechanism (e.g. brain), in order to actuate accordingly to the actual state of the agent. In order to understand the representation, we need to study both the physical structure, and the (functional) dynamics of the agent coupled with its environment (**Mitchell, 1998; Rocha, 2001**). Only a dynamical process can give a *value* to the physical structure. Different mechanisms can produce internal representations, but these are not *in* the mechanisms. We can say that the mechanisms are *external*, because we can analyze and decompose them. Representations of this type are *internal*, because they *depend* on the *relations* of the agent mechanism, sensors, and actuators, with its environment. We cannot take a representation "out of the agent", because it is "in" the relationship between agent and environment. They have a physical part, but we cannot understand it without the relationship. A physical state can represent more than one thing. What gives value to that state, what makes the state a representation, is the relationship.

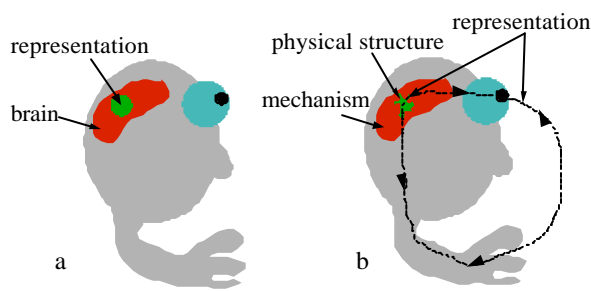

**Figure 1. Different views of internal representations**

We are interested in the development and internal dynamics of representations because the dynamics could be seen as the "motor" of a *closure process* within systems (**Pattee, 1995; Collier, 2000; Rocha, 2001; Emmeche, 2000**). The view of Maturana and Varela (**1987**) of organismic embodiment already assumes a complete closure. However, we are interested in how this closure comes to place. The closure process can be seen as a mode of description complementary to physical laws (**Pattee, 1995**). We are not interested on how an *action selection mechanism* can emerge depending on the predictive power of representations. We are interested in the *development* of representations, not in their *use*.

Representations need to be non-random, capturing some particular *structure* of the interaction with the environment. The more structure representations capture, the more "potential use" and "knowledge" they will have. In an epigenetic process, the formation of representations must catch first the simplest type of structure, i.e. historical embodiment. However, considering just this type of embodiment is not enough to provide any autonomy to the system. This is why we are interested in a simple system's "internal dynamics" helping to bootstrap more structure into the representation.

In this paper we examine how a *behavior modifier* (**Dörner, 2003; Dörner and Hille 1995**) affects the structural properties of an evolving representation as an internal dynamics responding to the *agent's* knowledge state. Behavior modifiers can produce behaviors which can be considered as emotions by an observer.

The representation will be considered as a network (directed graph) built under certain rules which compose a *closure mechanism*. This incorporates sensory signals and actuations, and relates and integrates them. The closure rules imprint by themselves some structural properties because they grow only by detecting and incorporating *affective states* (distinguishable states) as nodes (**Sheutz, 2000; Sheutz and Sloman, 2001**) and the relations between them (actuations) as arcs in a series of simple steps.

The reported experiments analyze two aspects: 1) the structural properties of the resulting representation as indicators of the agent's assimilated "knowledge" resulting from the interaction with the environment, being the useful macrovariables from an *objective* point of view; and 2) the closure mechanism's dynamics, as the internal, and therefore *subjective*, way the system develops its representation. The hypothesis is that modifying appropriately the behavior in function of its subjective "state of knowledge", the system will obtain benefits in the overall "macro" structure.

## 2. Methodology

To develop and analyze representations, we need conditions in which the agent copes with enough number of similar situations to "catch" (or "construct") knowledge as Piaget thought (**Montangero and Maurice-Naville, 1997**). The easiest way to do this is by interacting in an environment, simple enough to provide similar (but not identical) conditions. This is achieved playing "*pragmatic games*": every time an event occurs, the scenario restarts with similar conditions. This term is used here in resemblance of 'Language Games' (**Steels, 1996**), but as a methodology to study epigenetic development.

A characteristic of pragmatic games is that they can be carried out by the agent only by chance, as a result of its capabilities and the environment's characteristics. It means that the agent has the possibility to play the game and complete it with no more than the inborn capabilities. The agent can move allowing errors. As in language games, pragmatic games have no punishment or reward, success or failure.

## 3. Experimental setup

The implementation we developed to contrast our ideas is a pragmatic game called the "feed game", a subset of the micro-world used by Drescher (**1991**) to study Piaget's Schemes. This involves a 2D 7x7-grid world in which there is an agent consisting of one 5X5-grid "eye" with a central 1x1 square 'fovea', a 1x1 "hand", and a 1x1 "mouth". Within the world "objects" of size 1x1 can exist. The agent has four independent *actuators*, to move its hand and eye in the two dimensions. The eye's movements are restricted to focusing of the fovea within the world. The hand has the same constraint.

In the feed game, an object is placed randomly in the environment. If the hand passes over the object, it will be attached to the hand. If the hand holding the object passes over the mouth, the object is "eaten", and a new object appears at a random location, and the game starts again.

In Figure 2 we can appreciate a snapshot of the experimental setup: the grid stands for the visual field of the eye, with the fovea in the centre. A green (light) square at the bottom of the environment stands for the mouth. The hand is represented by a blue (dark) square, and the object by a red circle.

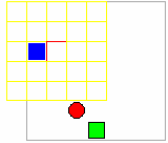

**Figure 2. Feed Game**

During the feed game, each of the four *actuators* chooses randomly among three possible options: decrease, maintain, or increase (-1, 0, or 1) the actual positions of the eye and hand in both dimensions (*ex*, *ey*, *hx*, *hy*), constrained by the environment. An *actuation* would be a set of four values of the *actuators*. An *actuator* is specified by a set of four (-1, 0, or 1) values ex, ey, hx, and hy. These are, respectively: the displacement for the eye in the x and y directions and the displacement for the hand in the x and y directions.

Each one of the 25 eye's cell senses the colors R (red), G (green), or B (blue) of the objects in the visual field, sending 3 bits to the agent, one for each color. Then the eye's sensing signal consists of 75 bits. The 'hand' and the 'mouth' contribute to the total *sensing vector* with one bit each one indicating if their position coincides with an object. Then a *sensing state* is conformed by a vector of 77 bits. The agent has no proprioception, in the sense that it has no register of the relative position of its hand, eye, nor mouth.

Additional to the *sensing states*, the system has other input signal from a set of distinguishable innate *biological motivations*. These are indicated by a 5-bit vector: Three bits for the fovea, each one for detecting R, G or B, one for the "hand" and other for the "mouth". Therefore, there are potentially 32 different biological motivations, although in our simple simulations less than ten are bootstrapped. These *biological motivations* do not have any 'appetitive' or 'aversive' character. They only are distinguishable and at the beginning they are not related with any *sensorial state*.

## 4. Closure Mechanism

Searching generality, we consider the use of *directed graphs*, a type of networks, as an adequate way to obtain and develop a representation for the agent.

We consider a *signal* as a situation which can be *distinguishable* for the agent and has been incorporated in the representation. In a simplified way, we will say that a process is *closed* if an *actuation* is related with the *signals* and the *signals* are related with the *actuation*. In this sense, the *closure mechanism* must be a process considering *how to introduce significant signals and actuations in the representation and how to identify if they are or not related*. In our directed graph, the nodes will have the *signal* information and the arcs the *actuation* information.

The dynamics and structural properties resulting in the network will be strongly related to the particular choice of the *closure mechanism* because this affects how nodes and arcs are introduced to the network **(Strogatz, 2001)**. The *closure mechanism* was thought in probabilistically favoring category's formation **(Hillman, 1997)**.

The *closure mechanism* will incorporate relevant nodes and arcs, modifying their status. It reaches a *class* of "well formed" links between nodes, they will be called *facts*, having some relation with the **(Drescher, 1991)**'s *schemas* but constructed with different criteria and motivations[1].

To develop the representation, we incorporate only nodes which are *affective states* **(Sheutz, 2000; Sheutz and Sloman, 2001)** for the agent. They represent any state of the agent that could affect it, for better or worse; including emotions, pains, desires, preferences, etc., but *not* measures. For our aim, we are only interested in their *distinguishable* character, being indispensable to filter information from the world and to establish organization. *Biological motivations* are considered as *affective states*, since they are distinguishable.

With this conception, *sensing states* are <u>not</u> *affective states*, because sensing alone has no relevance to the agent. The importance of a *sensing state* requires to be captured into a representation in order to acquire a *value* (relative to the agent). A *sensing state* can give place to an *affective state* if some *value* becomes associated to it. An *affective state* can be seen as a *signal*. These can become related through *actuations*. The possibility of establishing these relationships allows the agent to develop a structured representation in an autonomous way. The degree of integration of these relationships reflects the "knowledge" the agent has about its world. In our implementation, there is no use of this knowledge, but certainly it could be incorporated to perform e.g. some goal-directed task. Even when the *actuations* are random, the relationship between *signals* and *actuations* is structured. This structure is reflected in two ways: internally, during the *closure mechanism*, and externally, analyzing the network properties. Therefore, two modes of description are necessary **(Pattee, 1995)**.

In our experiments, the initial representation/network is empty. The agent has as inputs the *sensing states* and *biological motivations*. Every time the agent experiences a particular *biological motivation*, a record is created saving the *sensing states* associated with it.

A *sensing state* can give place to an *affective state* or a *potential affective state* –and then incorporated to the representation—by two mechanisms:

1. Detecting *affective states* from *biological motivations*. After a certain number of iterations (500 in our simulations), the system "falls" in a process in trying to determine if the *biological motivations* have some specific associated *sensing state*. Then the *affective state* represents the *sensing* bits always present when the *biological motivation* has been experienced.

2. *Potential affective states*. If the *sensing state* at time *t* corresponds to an *affective state*, then a node

---

[1] For Drescher, a casual (functional) relationship is searched, being reliability the last test. For us, only a significant departing from randomness in *actuations* establishes a deep relationship between affective states, stressing a non functional relationship between them.

corresponding to the *sensing state* in the time *t-1* is incorporated in the representation, as well the directed arc between the nodes (representing the *actuation*). The new node is called *potential affective state*.

A *potential affective state* becomes *affective state* if its frequency exceeds some value[2] (8 in our simulations).

The relationships between nodes (arcs) can be incorporated in the representation in two ways:

1. When a *potential affective state* is created.
2. When the agent experiences two sensing states having associated existing nodes (*affective states* or *potential affective states*) in the representation.

Every time an arc is crossed, the frequency and the performed *actuation* are recorded in the arc.

Once the arc exists in the network, its status can be modified with the recorded *actuations*' information during the process in the following way:

1. If the frequency of occurrence in experiencing a specific arc is larger than a given value (8 in our simulations), it becomes a *frequent arc* and the movement's distribution of probabilities for the *actuators* are computed from the history and saved in the arc in the form: {{p($ex$=-1), p($ex$=0), p($ex$=1)}, {p($ey$=-1), p($ey$=0), p($ey$=1)}, {p($hx$=-1), p($hx$=0), p($hx$=1)}, {p($hy$=-1), p($hy$=0), p($hy$=1)}}.

2. If in one of the 12 probabilities of a *frequent arc* is greater than a threshold (0.5 in our simulations), it is considered as a *codifiable arc*, since the nodes joining the arc have more than a random link, as the movement could be codified for at least one *actuator*.

If a *codifiable arc* has *affective states* as source and target nodes, it will be called a *fact*. This is considered the most refined state for the *closure mechanism* in the development of the arcs. This contrasts with the Drescher's perspective, which considers reliability as the way to verify the arc's functionality.

Our method is imperfect in associating nodes in a strict causal way, but has the advantage it has no intention into reach specific nodes or to proof specific arcs, avoiding any "cognitive" consideration. The representation is developed only with the (imperfect) agent's possibilities and not under our preconceived "true or false" considerations. This is because we are interested in the way the closure process occurs and not in its success as being "the best" fact constructor (Gershenson, 2004). Actually, further steps can be added to the closure mechanism defined here to refine the process.

However, we stress that the important aspect is the network formed by *facts*, and that the closure mechanism does not stop when an arc has achieved the *fact* character, but continuously incorporates new arcs and nodes (which can become facts).

---

[2] If the values are too small, noise can be learned. If the values are too big, then it takes more time to learn. This also happens for other parameters of the model.

## 4.1. Closure states

To identify the agent's closure state associated to every pair of consecutive *sensing states*, we use a set of three values (according to Table 1), one for the sensing in the time t-1, other for the sensing in the time t, and the third for the state of the link between them.

**Table 1. Codes of closure states**

| value | node | arc |
|---|---|---|
| 0 | not in representation | not in representation |
| 1 | potential affective state | not frequent |
| 2 | affective state | frequent |
| 3 | - | codifiable |

The closure state has 3x3x4 = 36 possibilities. For the specified *closure mechanism*, the state 223 has the highest "closure degree" and corresponds to a "closed" arc or *fact*. The particular closure's state is a *subjective* appreciation for the agent, in the sense that it does not tell anything to an observer who does not have precise knowledge of the mechanism.

Table 2 shows an example of a closure path towards forming a *fact*, following the states of nodes and arc 000 → 020 → 121 → 223.

**Table 2. Example of a closure path**

| Closure State | 000 | 020 | 121 | 223 |
|---|---|---|---|---|
| Var. focus | low | low | high | high |
| | a) | b) | c) | d) |

## 5. Behavior modulators

As mentioned before, we are interested in evolving structured representations beyond the possibilities that historical embodiment can offer. Following a bottom-up approach, we consider the importance of emotions (**Minsky, 1980**), being essential in the "independent dynamics" of representations. Still, we are interested in the emotional modulation of cognition (**Dörner and Hille, 1995; Dolan, 2002; Dörner, 2003**). In this sense, emotions are the observable result of a particular set of values for the behavior modulators.

We will consider a humble approach to Dörner's (**2003**) theory to test the idea that modifying the behavior according to the "closure state" (actual knowledge state) of an agent can help obtain more "knowledge". We use a single *behavior modulator* which we call *focus*. This parameter, with values between 0 and 1, modulates the probability to revisit the previous *sensing state*, by undoing the last performed movements. It is called *focus* because it is a mechanism used to perceive again something by trying (with a probabilistic measure) to revisit a situation.

A focus with value 0.0 means that the agent will always move in a random direction. A focus with value 1.0 means that the agent will always undo the last movement. A focus value of 0.5 will make the agent to move with the same probability than to undo the last movement. Different focus values can be interpreted by an observer as different emotional states. For example a high focus value could be seen as "interest".

## 6. Experiments

The experiments will be described from two perspectives, one *internal* corresponding to the closure mechanism's dynamics in incorporating signals and actuations, and other *external* corresponding to a network's quantifiers as being macro variables. With this we attempt to address the constrained problem posed by Pattee **(1995)** of how matter and symbol are related.

### 6.1. Internal Description

#### 6.1.1. Closure Dynamics: Constant Focus

In a first set of experiments, we perform runs of the *feed game* during 15000 iterations, each one with different focus value: 0.0, 0.25, 0.50, and 0.75. When focus has a value of 0.0, all the agent's movements are random. On the others, all the movements are random but having a probability –equal to the focus value– to 'undo' the last movement.

We are interested in the way the focus affects the representation's development given the system and the agent's closure mechanism. Every time the system changes the closure's state, either by incorporating a node, an arc, or changing their status in the representation, the agent has integrated more knowledge from the environment and its interrelationship, captured in the network.

In order to follow the closure's dynamics – knowledge's incorporation dynamics– a *probabilistic network* is built during the process considering the *closure states* as nodes and the possible *changes* between them as arcs, being weighted by the frequency of occurrence.

We distinguish between two types of arc: *loops* and *transitions*. Those situations in which the agent remains in the same *closure state* for consecutive iterations –i.e. without knowledge incorporation– will be called *loops* and represent time without knowledge acquisition. Table 3 shows the *loops'* relative frequencies for each focus (shown only =0.01 due to space restrictions). These represent the proportion of the global time the agent has experienced each type of *loop*. As the last row of the table shows, the agent is engaged in *loops* in more than the 85% of the total time. During loops, there is no distinguishable change in the closure mechanism, although there can be changes in the counters (e.g. it takes 8 occurrences to obtain a frequent arc). The number of *loops* changes according to the focus value; a lower *loop* frequency indicates less time without changes in the representation, learning faster. The focus plays a different role in learning depending on the specific loop. For example, in the loop 222-222 higher focus values are more convenient for the agent. On the other hand, low or no focus is convenient for the loop 000-000. There is no "best" focus value, but the "optimal" value depends on the actual (internal) context.

When the system experiences a change in the closure state, we can say that the system has incorporated structure in the representation. They will be called *transitions*. Table 4 shows the relative frequencies of transitions. The total time the *agent* develops its representation (*transitions*) is lower than the time devoted to *loops*.

#### 6.1.2. Taking Advantage of the Closure's Dynamics: Variable Focus

After analyzing the *probabilistic networks* corresponding to the *closure structures* obtained with each of the focus values, we chose for each *closure state* which focus value would be convenient to follow a path to the 223 state (*facts*) with less loops.

Figure 3 shows in the probabilistic network of *transitions* how the focus values were selected. The thickness of arrows indicates the transition probability. Darker arrows mean that for a higher focus there is a higher probability for *transitions*, while lighter arrows indicate lower probabilities while there is a high focus. Dotted lines do not contribute to the development of *facts*, and their width does not represent their probability.

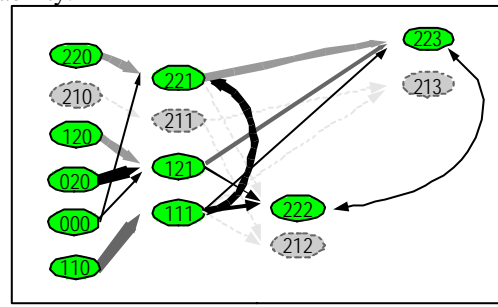

**Figure 3. Probabilistic network of closure dynamics**

We repeated the *feed game* experiment but changing the focus value in function of the actual *closure state* according to the following simple rule:

If {$closureState in [222 221 211 121 212 223 213]} {set focus 0.66} else {set focus 0.0}

The system changes its focus value in a reactive way, depending only on the current *closure state*, not on the *sensing states*.

The goal is not to find the "best" focus value for each *closure state*, but just to show that modifying the *behavior modulator* in terms of the *closure state* affects the acquired knowledge, obtaining a different structure of the global representation.

The results are given in the lasts columns of Table 3 and Table 4. We can see that the *loops* in which the agent spends more time decrease sensibly, and that the *transition* frequency rise or are maintained except for the 110-111 case. A detailed analysis of the data shows that the paths to the state 223 (*facts*) are favored.

The Average Time per Transition (ATT) is the average number of iterations in which the agent experiences a change in the *closure states*. It reflects the required time to incorporate in the representation a new aspect obtained from the interaction with its environment. Lower value means faster 'learning'. In Figure 4 we can see that a low focus enables the agent to learn faster than a high focus, but with a variable focus the agent learns even faster.

**Table 3. Relative frequencies of loops (=0.01)**

| loops | Focus value | | | | |
|---|---|---|---|---|---|
| | 0 | 0.250 | 0.5 | 0.75 | var |
| 222-222 | 0.31 | 0.26 | 0.25 | 0.23 | 0.17 |
| 000-000 | 0.12 | 0.15 | 0.16 | 0.24 | 0.09 |
| 223-223 | 0.09 | 0.09 | 0.12 | 0.14 | 0.19 |
| 221-221 | 0.08 | 0.06 | 0.07 | 0.08 | 0.14 |
| 121-121 | 0.05 | 0.06 | 0.06 | 0.05 | 0.09 |
| 111-111 | 0.05 | 0.05 | 0.04 | 0.03 | 0.01 |
| 211-211 | 0.05 | 0.05 | 0.05 | 0.04 | 0.07 |
| 100-100 | 0.04 | 0.05 | 0.05 | 0.04 | 0.03 |
| 010-010 | 0.04 | 0.05 | 0.05 | 0.04 | 0.02 |
| 200-200 | 0.03 | 0.02 | 0.02 | 0.01 | 0.02 |
| total | 0.86 | 0.85 | 0.87 | 0.91 | 0.83 |

**Table 4. Relative frequencies of transitions (=0.01)**

| transitions | Focus value | | | | |
|---|---|---|---|---|---|
| | 0 | 0.25 | 0.5 | 0.75 | var |
| 110-111 | 0.05 | 0.05 | 0.04 | 0.01 | 0.03 |
| 020-121 | 0.03 | 0.03 | 0.02 | 0.01 | 0.03 |
| 210-211 | 0.02 | 0.03 | 0.03 | 0.02 | 0.04 |
| 120-121 | 0.01 | 0.01 | 0.02 | 0.01 | 0.02 |
| 221-223 | 0.01 | 0.01 | 0.01 | 0.01 | 0.02 |
| 220-221 | 0.01 | 0.01 | 0.01 | 0.01 | 0.02 |
| 121-223 | 0.00 | 0.00 | 0.00 | 0.00 | 0.01 |
| total | 0.14 | 0.15 | 0.13 | 0.09 | 0.17 |

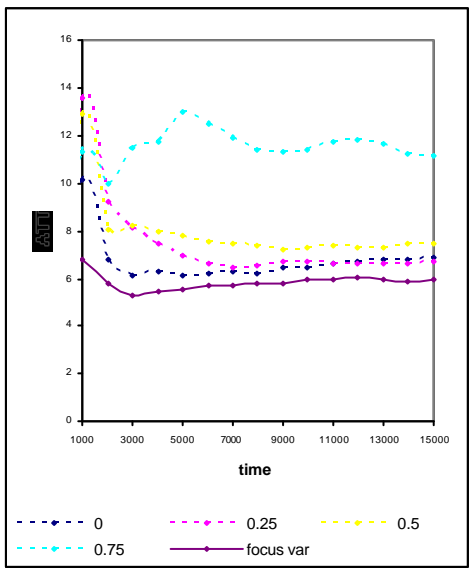

**Figure 4. Average Time per Transition**

## 6.2. External description: network properties

Both *loops* and *transitions* as considered before are characteristic of the *closure mechanism*, reflecting the dynamics during the development of the network. Their analysis can be seen as made from an *internal* mode of description, because an observer would not have access to these processes in an animal (i.e. symbols). We now will make an analysis from an *external* mode of description, "dissecting" the structure of the agent.

We can calculate the *closure state distribution* obtained from the resulting representation considering all the existing arcs and their associated nodes, as shown in Table 5. This distribution can be considered as an external observation, because it is a "picture" of the representation at a certain time, but does not give information on *how* the arcs have obtained their *closure state*. We can observe again that the focus affects the *closure state* of arcs in different ways. Although the "interesting" class of arcs is the 223, or *facts,* meaning the most elaborated kind of relation between signals and action. For *facts,* there is no significant variation with fixed focus values. But their frequency is increased in an important way (~50%) with variable focus. The trace on how *facts* are favored in this case can be followed in time in Figure 5.

We used the Clustering Coefficient **(Watts and Strogatz, 1998)** of the representation network as a measure of global structure. In Figure 6 the Clustering Coefficient shows that the more efficient *and* more stable case is the one of variable focus. Both figures indicate that having a variable focus yields a high "volume" with "appropriate" density in the representation network. The variable behavior modulator actuates *internally* to produce improvements in the *external* structure. In our model, the sensed turns into signal *internally* through the closure process, and the result can be measured *externally* in the structural properties of the representation network.

**Table 5. Closure state of arcs of final representation**

| arcs | Focus | | | | |
|---|---|---|---|---|---|
| | 0 | 0.25 | 0.5 | 0.75 | var |
| 121 | 586 | 539 | 503 | 243 | 586 |
| 221 | 386 | 338 | 271 | 272 | 409 |
| 211 | 264 | 341 | 364 | 191 | 447 |
| Facts: 223 | 213 | 188 | 204 | 211 | 307 |
| 111 | 370 | 477 | 337 | 97 | 196 |
| 222 | 59 | 59 | 43 | 43 | 94 |
| 213 | 6 | 1 | 8 | 4 | 5 |
| 212 | 1 | 1 | 2 | 0 | 0 |
| num.arcs | 1885 | 1944 | 1732 | 1061 | 2044 |

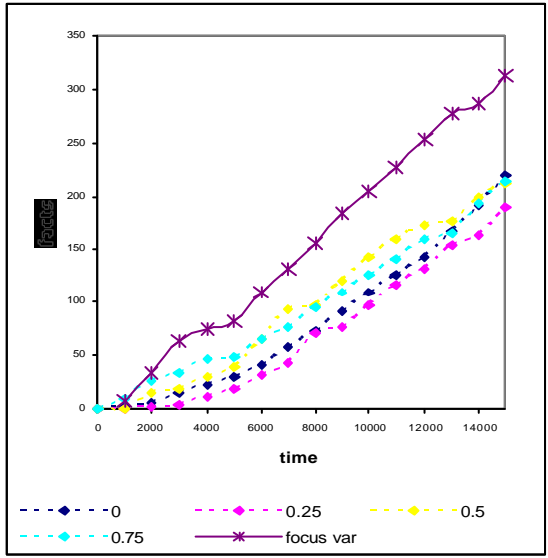

**Figure 5. Facts for Feed Game**

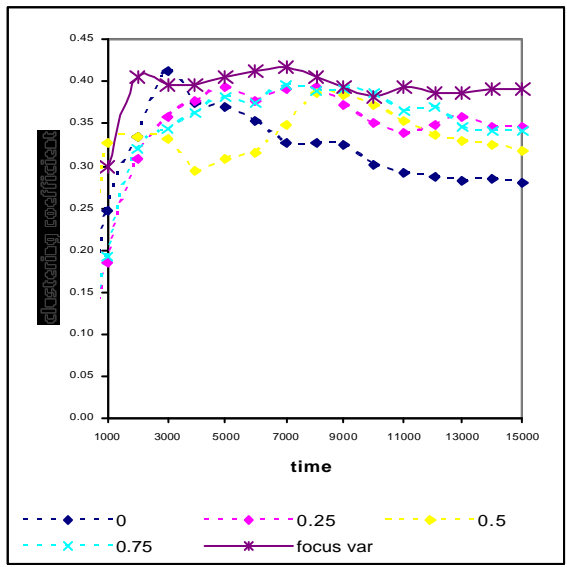

**Figure 6. Clustering coefficients for Feed Game**

We also analyzed the subnets related to each *biological motivation* (data not shown). Each subnet is similar to a scheme, more in the Piagetian sense than in the Drescher's sense. This is because the subnet corresponds to structured knowledge with some biological meaning and not to concrete context-actuation-result detection. The structural properties of subnets are better with a variable *focus* than with a fixed one. There are also more subnets with a variable *focus*, giving the possibility to develop more schemes.

## *7. Conclusions and Future Work*

In this work, we isolated a representation's evolving process, avoiding any use of "cognition" about the "state of the world", to explore two aspects: a) the necessity of two modes of description in the Pattee's sense, and more simply, a case study on how the internal and external modes of description can be related, and b) how behavior modifiers based on internal considerations can affect the structural properties of the developed representation

a) For the internal mode of description, we consider a built in mechanism called *closure mechanism* determining how and when to incorporate structurally related *signals* and *actuations* in a representation. The *signals* are assigned in terms of the *sensing states* but only when they are related directly or indirectly with a biological motivation. The *actuations* are incorporated in terms of *actuator*'s movements. The structural relationships are given in different stages, until possibly reaching the most elaborated state or *fact*: when a couple of adjacent *affective states* can be related more than in a random way. The *closure mechanism* has its own dynamics and can be observed as a probabilistic network which forms *facts*, being the internal mode of description. The external mode of description is given in terms of the properties of the network of *facts*.

b) The *closure mechanism* is in itself a "knowledge acquisition mechanism" in the sense that it incorporates in the representation the structural relationships with the environment. Using a behavior modulator called *focus*, the representation and its structure can develop in different ways. A selective value (i.e. variable focus) for specific *closure states* improves the structural properties of the representation (external mode of description).

Our model is not a behavior-based or knowledge-based, but as simple as a reactive system. It is atypical for a "knowledge acquisition mechanism", since the agent does not react to its world. However, the *focus* can modify the behavior patterns. The variable *focus* allows the agent to react to its knowledge state in order to incorporate faster the relationships with its environment. The obtained representation does not catch "structural" properties of the environment, but makes explicit the structural *interactions* between the agent and environment. We have avoided any use of the representation but these have a potentiality to be used.

Only structural or historical embodiment is not enough for obtaining autonomously rich representations. It seems the same as to think about only affordances. We need consider also an internal process "independent to the world's dynamics" **(Steels, 1995)**, in such a way that the representation becomes richer. Note that the mentioned dynamics is different from the related with the *use* of representations. In the agent's life, both dynamics are crucial, and must be related, but at this moment we are concentrated in building representations.

We conclude:

a) Considering only dynamical aspects of the system-environment interaction can give us only historical embodiment. To obtain more structured representations we need to explore internal mechanisms to understand how structural properties can be bootstrapped.

b) Structural properties in developing representations can be changed by using *behavior modifiers*, not considered as the nervous system in the epigenetic process but as the –less often considered but more basic– endocrine system which is related to emotions.

c) A selective use of this parameter depending on the *closure state,* ("knowledge state"), improves the bootstrapping of structure. We consider this as a fundamental step in understanding how the use of representations can rise to manage "knowledge" starting from reactive systems.

d) The resulting system with a simple "internal dynamics" is a *knowledge acquisition mechanism* allowing more structure in the resulting representation that the solely historical or structural embodiment can provide but without any cognitive, purposeful or intentional consideration.

e) For studying representations, both a structural and a dynamical description are necessary. This helps in elucidating the problem of how symbols are related to matter (**Pattee, 1995**).

As a future work, we can see several directions which could be followed. Intersubjective representations could be obtained by pragmatic games in which two or more agents interact with an environment. This topic is interesting for studies in the evolution of communication. Another direction would be to study the effect of different *behavior modifiers* in the development of the representation.

## *Acknowledgements*

C. R. de la M. B. was supported by CONACyT, México and Universidad Veracruzana. C. G. was partially supported by CONACyT, México.